\documentclass[runningheads]{llncs}
\usepackage{graphicx}
\usepackage{cite}
\usepackage{amsmath,amssymb,amsfonts}
\usepackage{algorithm}
\usepackage{algorithmic}
\usepackage{graphicx}
\usepackage{textcomp}
\usepackage{xcolor}
\usepackage{array}
\usepackage[caption=false,font=normalsize,labelfont=sf,textfont=sf]{subfig}
\usepackage{stfloats}
\usepackage{url}
\usepackage{verbatim}
\usepackage{color}

\usepackage{algorithm}
\usepackage{stmaryrd}
\usepackage[a4paper, total={184mm,239mm}]{geometry}
\usepackage{listings}
\begin{document}
\title{ECSAS: Exploring Critical Scenarios from Action Sequence in Autonomous Driving}
%
%\titlerunning{Abbreviated paper title}
% If the paper title is too long for the running head, you can set
% an abbreviated paper title here
%
\author{Shuting Kang\inst{1,2} \and
Heng Guo\inst{1,2} \and
Lijun Zhang\inst{2} \and Guangzhen Liu\inst{2} \and Yunzhi Xue\inst{2} \and Yanjun Wu\inst{2}}
%
% \authorrunning{F. Author et al.}
% First names are abbreviated in the running head.
% If there are more than two authors, 'et al.' is used.
%
\institute{University of Chinese Academy of Sciences, Beijing, China \and  Institute of Software Chinese Academy of Sciences, Beijing, China \\
\email{kangshuting18@mails.ucas.edu.cn, guoheng21@mails.ucas.edu.cn}\\
%\url{http://www.springer.com/gp/computer-science/lncs}
}
\maketitle              % typeset the header of the contribution
%

% \begin{figure}[H]
% \hsize=\textwidth
%  \centering
%  \setlength{\abovecaptionskip}{0cm} 
%         \includegraphics[width=\linewidth]{figure/motivation.PNG}        
%         \caption{An illustration of our motivation.  Compared with past studies sampling from all parameters simultaneously, our study samples  based on the order of parameter.  }      
%  \label{motivation}
% \end{figure}

  % is indispensable for the effect of scenario generation.

\begin{abstract}
Critical scenario generation requires the ability of sampling critical combinations from the infinite parameter space in the logic scenario. 
Existing solutions aim to explore the correlation of action parameters in the initial scenario rather than action sequences.
How to model action sequences so that one can further consider the effects of different action parameters in the scenario is the bottleneck of the problem.
In this paper, we attack the problem by proposing the ECSAS framework. Specifically, we first propose a description language, BTScenario, allowing us to model action sequences of the scenarios. We then use reinforcement learning to search for combinations of critical action parameters. To increase efficiency, we further propose several optimizations, including action masking and replay buffer. We have implemented ECSAS, and experimental results show that it is more efficient than native approaches such as random and combination testing in various nontrivial scenarios.

\end{abstract}
\section{Introduction}
\begin{figure}[htbp]
    \centering
     \includegraphics[width=\textwidth]{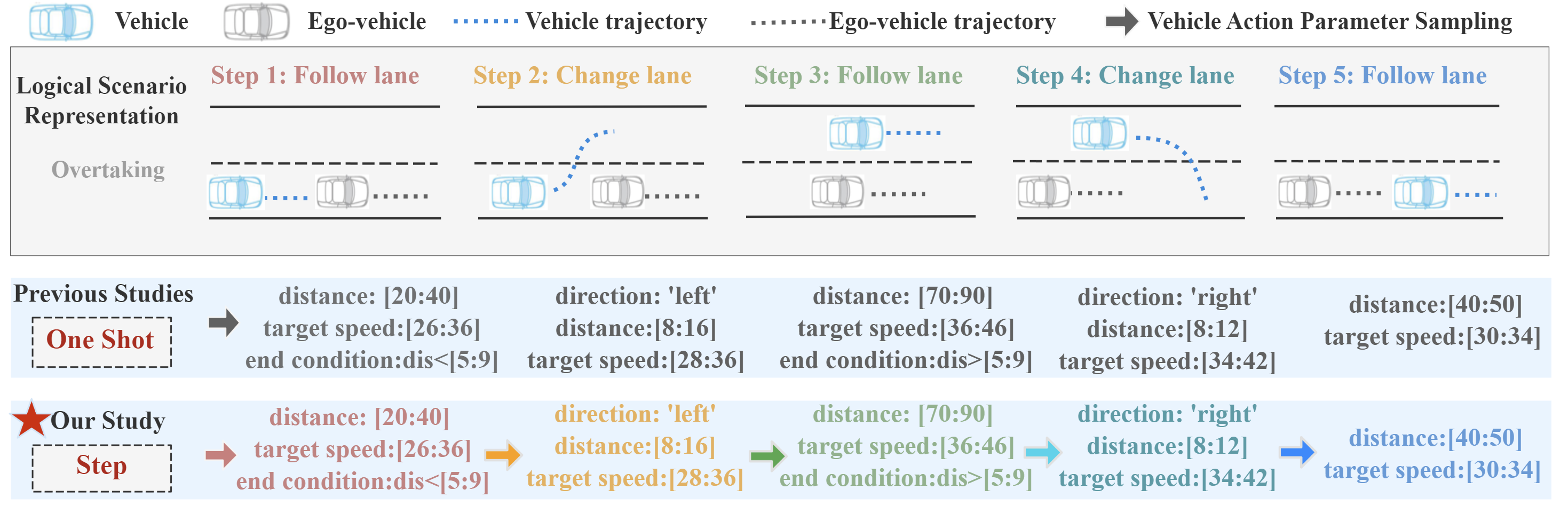}
	\caption{An illustration of our motivation. Compared with past studies sampling from all parameters of a logical scenario at once, our study generates action parameters by step from the perspective of action sequence. Subsequent experiments also verified this. }
    \label{motivation}
\end{figure}

Despite the great success in \emph{Autonomous Driving Systems} (ADS), safety engineering is still challenging for complex and dynamic environments, also known as scenarios. The space of possible test scenarios is virtually continuous, it is thus a challenging task how critical scenarios are generated. With critical scenarios, we refer to situations that cause potential safety risks \cite{zhang2021finding}. \emph{Critical Scenario Generation}  (CSG) aims to find concrete critical scenarios within a given logical scenario, where the logical scenario is on a  state-space level with parameter ranges, while a concrete scenario is a concretization of a logical scenario with concrete parameters \cite{menzel2018scenarios}. Therefore, how to find critical parameters from infinite space becomes a core challenge for CSG.

To advocate research in this direction, native approaches, such as random testing (RT) and combination testing (CT), sample directly from parameter ranges of actions. These  approaches are inefficient when critical scenarios are rare. To attack this problem,  recent works \cite{ding2020learning,ding2021multimodal,ding2021causalaf}  focus on exploiting the correlation of actor's parameters by applying reinforcement learning (RL). However, they compute the probability distribution of all parameters in the scenario from the beginning. 

Actually, a scenario is often modeled as a sequence of atomic actions, and the  temporal relation between action parameters should also be considered. 
In this paper, we propose the \emph{ECSAS framework}, abbreviating for \emph{\textbf{E}xploring \textbf{C}ritical \textbf{S}cenarios from \textbf{A}ction \textbf{s}equence}. The core mechanism behind ECSAS is to model the action sequence of each scenario, so as to consider the influence relation between the action parameters. 
Considering the example in Fig. \ref{motivation}, we model an overtaking scenario with two atomic actions (follow lane and change lane) in five steps.
Previous works sample all action parameters simultaneously in the overtaking scenario at the start. 
Assume that we have  sampled values of the first step. In our framework, decision of the second step will make use of the values from previous steps, allowing ECSAS to search for critical scenarios efficiently. 

% ,openscenario1.0,openscenario2.0

For modeling scenarios,  scenario description languages provide rich APIs, which are  more flexible and convenient than the graphic model (Bayesian probability graph \cite{ding2020learning}, causal graph and behavioral graph \cite{ding2021causalaf}), temporal logic \cite{gladisch2019experience}, and Markov decision model \cite{wachi2019failure, koren2019efficient, koren2018adaptive}.
Current description languages can be divided into two categories: representing scenarios from a local viewpoint \cite{scenic,scenic2,paracosm}, or using the positional changes from a global viewpoint \cite{GeoScenario}. However, these works fail to model scenarios based on the  action interactions. Thus, in this paper, we propose a new language, BTScenario, which supports the representation of action sequences using temporal operators  (``serial'' and ``parallel'').  

Inspired by \cite{ding2020learning}, we use reinforcement learning (RL) to synthesize critical parameter valuations in logical scenarios. Action sequences leading to collisions are assigned with higher rewards, then the problem reduces to synthesizing optimal policies of the model. We use TD3 for this purpose, one of the RL algorithms to handle the continuous parameter spaces. We use an actor network to select actions and two critical networks to evaluate action sequences. During training, we use the action mask to fix the length of action space and the order of action, which means that the choice of next-step parameters does not affect the state of the previous step. We also optimize the storage mechanism, replay buffer, to improve search efficiency. 

Summarizing, the main contribution of the paper is the  ECSAS framework for generating critical scenarios. We propose  a scenario description language, BTScenario, to model the action sequences in logical scenarios. ECSAS then uses RL combined with the Carla simulator to synthesize critical scenario valuations. 
We have implemented the ECSAS framework. Comprehensive and extensive experiments are conducted, showing  that ECSAS outperforms native RT and CT approaches in various scenarios.

%Good performance can be well explained by focusing on the relationship of parameters between action sequences.  
% The main contributions of this work are as follows:
% \begin{itemize}
%  \item
% \item
% We design a scenario description language, BTScenario, to model logical scenarios. BTScenario supports an abstract and explicit specification of the interactions between actors in chronological order.
% \item  
% \end{itemize}

\section{Related Work}
\subsection{Scenario representation}
Scenario representation is the first step in critical scenario generation.
Let us briefly summarise Scenic 1.0 \cite{scenic}/2.0 \cite{scenic2}, Paracosm \cite{paracosm}, and GeoScenario \cite{GeoScenario} about scenario specifications in the sequel. 

% and OpenScenario 1.0\cite{openscenario1.0}/2.0\cite{openscenario2.0}

Scenic 1.0 focuses on the spatial layouts of map objects and the actors but does not allow specifying the temporal actions of actors. Scenic 1.0 was extended to Scenic 2.0 to remedy this deficiency. Scenic 2.0 specifies the actions locally in the sense that actions are specified separately for each actor. ``ego = Car with behavior EgoBehavior()'' is a simplified example to show this local viewpoint. 
Paracosm further abstracts the Python API interfaces of the simulator for users to support the action of actors in a way similar to Scenic 2.0.
% pedestrian = Pedestrian with behavior CrossingBehavior()

% In particular, it assumes a local viewpoint as Scenic 2.0.

% We also provide a simplified example as follows.
% \begin{lstlisting}
% v = AutonomousVehicle(loc = r.onLane (1,10))
% p = Pedestrian (start = r.onLane(r.SideWalk,100),target = r.onLane (-r.SideWalk ,100), dist = Const(30))
% \end{lstlisting}

Unlike Scenic and Paracosm, GeoScenario focuses on describing the scenario from a global viewpoint.  Based on XML, GeoScenario specifies the scenarios by describing the trajectories of actors by a sequence of nodes specified on the map, the triggers for starting the actors, as well as the expected speeds when reaching the nodes. 

% OpenScenario 1.0 defines the storyboard to describe the temporal actions of actors and events to trigger the start or end condition of the scenario. Nevertheless, the XML-based scenario description language exhibits complex formats and poor readability. In view of this, OpenScenario 2.0 introduces the serial and parallel operators to compose actors' actions by Python-like syntax.  OpenScenario 2.0 defines the position, speed, and other modifiers used to modify the action ``drive''.
% For instance, we use a lane change scenario to show interactions between actors. 
% \begin{lstlisting}
% do parallel():
%   v2.drive(p)
%   serial:
%    A: v1.drive(p) with:
%     lane(same_as: v2, at: start)
%     lane(left_of: v2, at: end)
%     position([10..20]m, behind: v2, at: start)
% \end{lstlisting}
% However, GeoScenario and OpenScenario simulate actions through positional changes, hindering the intuitive representation of the action interactions. 
However, both local view methods based on event-driven behavior patterns and global description methods based on trajectory points are not easy to split into the action sequences we want. 
We propose the BTScenario to model action sequences using temporal operators(``serial'' and ``parallel'') with encapsulated atomic action interfaces (e.g., ``followLane'' and ``changeLane'').

\subsection{Critical Scenario Generation}
Given a logical scenario, we aim to generate a set of concrete critical scenarios. 
A simple one-action scenario, such as going straight, can be sampled using mathematical formula constraints, but for complex multi-action scenarios, collision scenarios are most likely to occur during the first action phase, and user-defined mathematical calculations are required. Moreover, since the vehicle dynamic model in the simulation makes the vehicle no longer in an ideal driving state, it becomes impossible to rely solely on mathematical calculations.

A native scenario exploration approach, RT, randomly assigns each parameter's value in the logical scenario space. The CT \cite{kuhn2015combinatorial} aims to generate a minimum set of test cases that satisfy N-wise coverage. The CT can be used to find unknown combinations of parameters that may fail the ADS. However, these native approaches (RT and CT) can be inefficient because of the infinite parameter space and the low fraction of critical scenarios.

For the above challenge, the search-based  methods have the potential to be more efficient \cite{scenic,scenic2,ding2020learning,ding2021multimodal}  since the searching direction at each iteration is adjusted. However, gradient-based methods \cite{scenic,scenic2} only consider the role of a single parameter in CSG. 
Considering the correlation between parameters, most recent works sample critical parameters from a joint probability distribution of actor parameters by RL \cite{ding2020learning, ding2021multimodal} in a logical scenario. Another solution \cite{ding2021causalaf} introduces prior human knowledge by the causal graph and then represents the interaction by the behavioral graph. These methods mainly consider the correlation of all actor parameters from the scenario initialization. Instead, ECSAS considers the correlation of parameters in the action sequence, to the best of our knowledge, is the first framework for finding critical scenarios that adjusts decisions based on previous actions.

\section{Methodology}

 \begin{figure*}[tbp]
     \includegraphics[width=\linewidth]{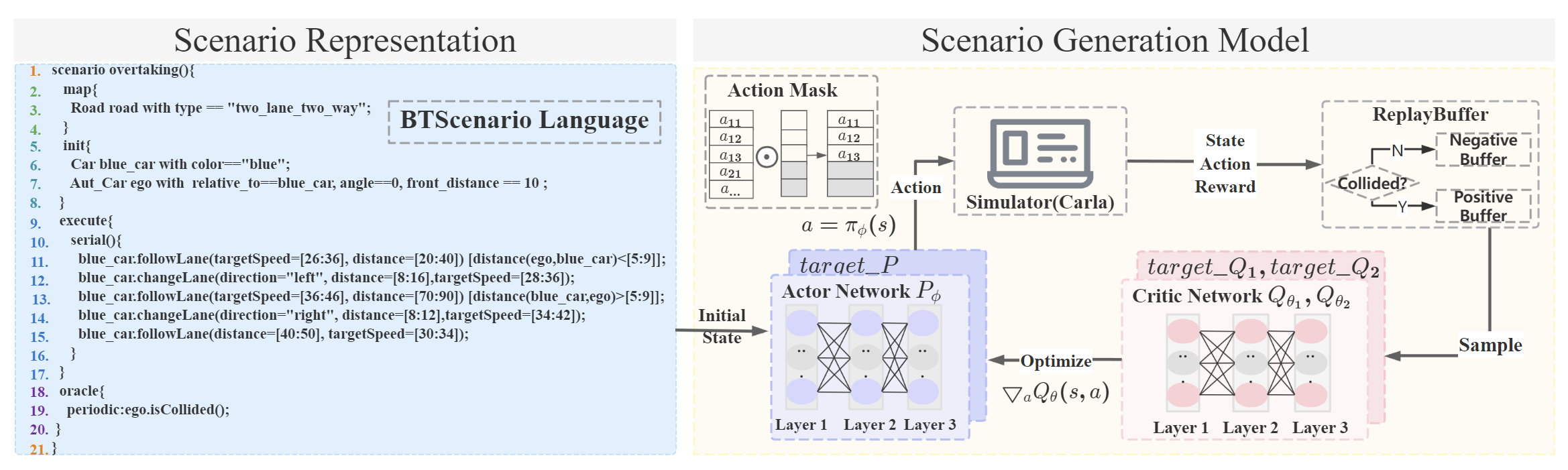}
	\caption{An overview of ECSAS. The model contains two modules: Scenario Representation aims to depict a logical scenario by BTScenario and ego-vehicle system, and Scenario Generation Model aims to explore critical parameter combinations by RL and generates concrete scenarios by the simulator.
    }
    \label{framework}
    %\vspace{-15pt}
\end{figure*}

%Given a logic scenario,  the CSG aims to find concrete critical scenarios as many as possible, which are collided in this paper.  The key is to find the critical parameter combinations in the parameter space of actions, i.e., $A={A_1,A_2,...,A_T}$, where $A$ is the set of actions' parameters of a scenario,  $T$ is the number of steps in the scenario, and $A_t$ contains the parameters of $N$ actions ${a_{t1}, a_{t2}, ..., a_{tN}}$ in the step $t$.  

%\textbf{LZ:Check this?}

A logic scenario is a sequence of actions' parameters $A={A_1,A_2,...,A_T}$, where $T$ is the number of steps in the scenario, and $A_t$ contains the parameters of $N_t$ actions ${a_{t_1}, a_{t_2}, ..., a_{t_{N_t}}}$ in the step $t$. The CSG aims to find \emph{concrete critical scenarios}, which are colliding scenarios in this paper.  Intuitively, concrete critical scenarios instantiate all parameters with concrete values so that the scenario leads to collisions.

Fig. \ref{framework}  illustrates the overall structure of ECSAS. We first propose a scenario description language, \emph{BTScenario}, allowing us to model the sequence of actor actions conveniently. Taking the BTScenario as an input, the  scenario generation model exploits RL to explore the critical parameters. Sampled values are fed to the simulator, which simulates the physical environment, returns the reward, and updates the state. By stacking the above two processes multiple times, ECSAS learns the actor and critic networks that encode concrete critical parameters.

\subsection{BTScenario: Scenario Representation Language}
Researchers have proposed various scenario definitions \cite{ulbrich2015defining,bock2018data,SWTL+21}, among which some common components exist: namely map, actor, interaction, and oracle. 
Inspired by these definitions, we propose BTScenario to model action sequences.
It offers rich map elements to describe the driving area (\ref{subsub:map}),  the specification of the initial state (\ref{subsub:initial}), interaction (\ref{subsub:tem}), and the test oracle (\ref{subsub:oracle}).
We design an overtaking scenario (see Fig. \ref{framework}) to illustrate some selected key map elements defined in BTScenario.
% \subsection{\textbf{Verification Experiment}}
% We give an overtaking scenario to show the generation of risky scenarios in Fig. \ref{motivation}. First, we specify the logic scenario by BTScenario as follows:
% \begin{lstlisting}[morekeywords={scenario, map, init, execute, oracle}]
% scenario overtaking(){
%   map{
%    Road road with "two_lane_two_way";
%   }
%   init{
%    Car blue_car with color=="blue";
%    Aut_Car ego with  relative_to==blue_car, angle==0, front_distance == 10 ;
%   }
%   execute{
%    serial(){
%     blue_car.followLane(targetSpeed=[26:36], distance=[20:40]) [distance(ego,blue_car)<[5:9]];
%     blue_car.changeLane(direction="left", distance=[8:16],targetSpeed=[28:36]);
%     blue_car.followLane(targetSpeed=[36:46], distance=[70:90]) [distance(blue_car,ego)>[5:9]];
%     blue_car.changeLane(direction="right", distance=[8:12],targetSpeed=[34:42]);
%     blue_car.followLane(distance=[40:50], targetSpeed=[30:34]);
%    }
%   }
%   oracle{	 
%     periodic:ego.isCollided()||blue_car.isCollided();
%   }
%  }
% }
% \end{lstlisting}
% When the scenario initializes,  the ego is 10 meters in front of the blue car. 
% In the ``oracle'' module, we consider the critical scenario referring to the ego or blue car crashing.

\subsubsection{\textbf{The Map}}
\label{subsub:map}
Lines 2-4, of the code in Fig.\ref{framework},  show how map objects are specified in BTScenario: a road named ``road'' is declared as "two\_lane\_two\_way".
The following code declares a junction named ``crossroad'' with type "+".
 \begin{lstlisting} 
 Junction  crossroad with type == "+";
 \end{lstlisting}
 Note that a junction can also be of ``T'', ``Y'', or ``unknown'' type, where the ``unknown''   type denotes the junctions, not of a fixed shape,  i.e. not of type ``+'', ``T'',  or ``Y''. 

Referring to the Apollo map structure \cite{apollo}, BTScenario supports eleven map objects(e.g., junction, road, lane, etc.), where several roads can be attached to a junction, and a road can have several lanes.

\subsubsection{\textbf{Initial States Of Actors}}
\label{subsub:initial}
In BTScenario, we consider various types of actors: \textit{Aut\_Car} represents the car under test (line 7  declares ``ego'' as the car under test), \textit{Car} represents the environment car (the ``blue\_car'' at line 6). Moreover, we can specify  additional constraints when declaring an actor. For instance,  line 6 specifies the car with parameter color defined to be ''blue''.

An important attribute of actors is their locations. BTScenario supports the declaration of both absolute  and relative locations. For instance, line 6 states that ``blue\_car'' is at a randomly chosen location of ``road'' by default; line 7 represents  ``ego'' is 10 meters ahead of ``blue\_car'', and ``angle==0'' represents both cars are in the same direction. 

% The following code states that  ``car2'' is at coordinate point (1,2), and ``car3'' is 100 meters ahead of ``car1''. 
% \begin{lstlisting}
% Car car1 with absolute_position == lane1;
% Car car2 with absolute_position == 1@2;
% Car car3 with relative_to == car1, front_distance == 100;
% \end{lstlisting}

\subsubsection{\textbf{Temporal Actions of Actors}}
\label{subsub:tem}
Before introducing the temporal actions of actors, we first show the specification of the atomic actions.
For cars, the atomic actions include \textit{followLane} and \textit{changeLane}, etc. 
BTScenario supports the two types of parameter expressions of actions. For example, line 11 specifies that the driving distance of ``blue\_car'' can be defined as a value within $[20:40]$.
Moreover, actions can have pre-conditions and post-conditions that specify the enabling and terminating conditions of the action.
For instance, line 11 specifies that ``blue\_car'' follows the current lane until the distance between ``ego'' and ``blue\_car'' is less than five to nine meters.

BTScenario specifies the interaction of actors  by two temporal operators: the \textit{serial} and \textit{parallel} operators, meaning the sequential and parallel composition of actions, respectively. 
Fig. \ref{framework} gives an overtaking scenario using \textit{serial}. Multiple  cars can run in parallel, illustrated by the following code:
 \begin{lstlisting}
     parallel(){
      car1.followLane(targetSpeed=[10,20]);
      car2.followLane(targetSpeed=[10,20]);
     }
 \end{lstlisting}

% Besides, we provide another case that ``car1'' and ``car2'' first follow the current lane in parallel,  then ``car1'' changes to the left lane, and ``car2'' keeps straight. 
% \begin{lstlisting}
% serial(){
%     parallel(){
%      car1.followLane(targetSpeed=[10,20]);
%      car2.followLane(targetSpeed=[10,20]);
%     }
%     parallel{
%      car1.changeLane(direction="left");
%      car2.followLane(targetSpeed=20, distance=30);
%     }
% }
% \end{lstlisting}

\subsubsection{\textbf{The Oracle}}
\label{subsub:oracle}
To specify the test standards for critical scenarios, there are two kinds of operators in oracle module of BTScenario: ``periodic'' and ``record''.  The operator ``periodic'' requires that the oracle needs to be verified unless it is violated or the simulation is completed.  The  operator ``record'' indicates that the program must record information about the actors or environment in the scenario. Line 19 specifies that the simulator monitors whether the ``ego'' collides.

\subsection{Scenario Generation Model}
% We use BTScenario to model the action sequences in the scenario. 

Because the iterative state-parameter selection process of CSG has some flavour of the \emph{policy iteration} of Markov Decision Process (MDP) \cite{abakuks1987reviewed},  we use RL, which is commonly used in analyzing MDPs, to explore critical parameter combinations between different steps.

A scenario $S$ consists of $T$ steps according to the temporal relation of actions. At each step $t\in\{1,\ldots, T\}$, with a given state $s_t \in S$, we select the concrete values of $A_t$ to generate  concrete scenarios.
In RL, we regard the environment vehicle as an agent. Below, we introduce our reward design, action mask, replay buffer optimizations, and the optimization process. 

\subsubsection{\textbf{Replay Buffer Optimization}}
Due to the scarcity of critical scenarios, a long-tail problem remains in the training process of CSG network\cite{ding2020learning}. To address this problem, we optimize the replay buffer of RL. We separated the buffer into two parts to divide the samples according to whether or not the vehicle collided in the episode. The samples are stored in the temporary buffer $R_{tmp}$ during an episode. After the episode is finished, the samples in $R_{tmp}$ are moved to positive buffer $R_{+}$ if the vehicle collides otherwise to negative buffer $R_{-}$. 
In training the ECSAS, we sample $\eta \cdot N$ samples from $R_{+}$ and $(1-\eta) \cdot N$ samples from $R_{-}$, where $0 \leq \eta \leq 1$.

\subsubsection{\textbf{Optimization Process}}
\label{OptimizationProcess}
Since TD3 \cite{fujimoto2018addressing}, one of the RL algorithms, can be used to handle continuous action parameter spaces and is insensitive to hyper-parameter settings, we follow it to solve our optimization problem. 

In RL, the objective is to find the optimal policy $\pi_\phi$, with parameters $\phi$, which maximizes the expected return $J(\phi)$. The optimal policy $\pi_\phi$ in ECSAS encodes the critical parameters of action $A$ causing the collision. For the continuous action space, $\pi_\phi$ can be updated by the gradient of the expected return $\nabla_\phi J(\phi)$. In TD3, the policy, known as the actor, can be updated by the deterministic policy gradient \cite{silver2014deterministic}:

\begin{equation}
    \begin{aligned}
        {
            \begin{matrix}
                \nabla_{\phi} J(\phi) =
                % \mathbb{E}_{s\sim p_{\pi}} [ \nabla_a Q^\pi(s,a)|_{a=\pi_\phi (s)}\nabla_{\phi} \pi_{\phi} (s)   ] \approx

                % \mathbb{E}_{s\sim p_{\pi}} [ \nabla_a Q_\theta(s,a)|_{a=\pi_\phi (s)}\nabla_{\phi} \pi_{\phi} (s)   ] \approx
            
                 N^{-1}\sum \nabla_a Q_{\theta_1}(s,a)|_{a=\pi_\phi (s)}\nabla_{\phi} \pi_{\phi} (s)
            \end{matrix}
        }
    \end{aligned}
\end{equation}
where $Q_{\theta}(s,a)$ is the action-value function, and the parameter values of actions $a \in A$ are generated from an actor network.
The critic network is updated by temporal difference learning to maintain a fixed objective $y = r + \gamma min_{i=1,2}Q_{\theta_i}'(s',\tilde{a})$ over multiple updates. The gradient for updating critic networks parameters $\theta$ is:
   \begin{equation}
    \begin{aligned}
        {
          \theta_i  = arg min_{\theta_i}N^{-1}\sum (y-Q_{\theta_i}(s,a))^2
        }
    \end{aligned}
    \end{equation}

\section{Experiment}

\subsection{Experiment Settings}
\begin{table}[tbp]
    \centering
     \caption{Hyper-parameters in experiments.}
    \begin{tabular}{c|c|c}
    \hline
    Hyper-parameter   &Description  &Value \\
    \hline
    $d$  &delayed steps for updating network	&3\\
     \hline
     $C$  &collision  reward	&5\\
      \hline
     $E$  &episode	&121\\
     \hline
     $N$  &batch size	&50\\
      \hline
    $\sigma$  &range of action noise	&[0,1]\\
     \hline
     $\tau$  &delayed update parameters	&1e-2\\
     \hline
    $\gamma$  &threshold in $Q$	&0.9\\   
     \hline
    $\eta$  & sample coefficient between $R_+$ and $R_-$ &0.5\\
     \hline
     \end{tabular}
     \label{table1}
\end{table}

The map data is a virtual town on the plain in Apollo format. In our implementation, we connect Carla as the back-end simulator and Unreal Engine4  as the engine. We apply Lark\footnote{https://lark-parser.readthedocs.io/en/latest/grammar.html} to compile the language and take PyTrees\footnote{https://apollo.auto/index\_cn.html} to construct and execute the behavior tree. We use the feed-forward PID speed control algorithm e \cite{gene2015feedback}, and the Stanley direction control algorithm \cite{hoffmann2007autonomous} for the ego car. Other details about the hyper-parameters in the Algorithm \ref{alg1} are listed in Table \ref{table1}.

\textbf{Metrics:} We follow the metric, collision rate (CR) \cite{ding2021multimodal,ding2020learning,ding2021causalaf,zhang2021finding} to evaluate the performance after a static policy of the model. It is the percentage of scenarios for which the ECSAS yielded a collision while respecting behavioral constraints. In particular, a scenario is only considered successfully collide if all environment cars do not collide with other environment cars.
CR is formulated in Eq.~\ref{cr_eq}, where $U$ refers to the number of successful collision scenarios, $W$ refers to the total number of tested scenarios.
\begin{equation}
    \label{cr_eq}
    \begin{aligned}{
          CR = \frac{U}{W} \cdot 100\%
    }\end{aligned}
\end{equation}

% \footnote{https://py-trees.readthedocs.io/en/release-2.1.x/index.html}
% \footnote{https://docs.unrealengine.com/en-US/index.html}
% 
% \subsubsection{\textbf{Evaluation Metric}}

 % and the number of iterations required for the model to reach stability

\begin{figure}[tbp]
 \centering
     \includegraphics[width=0.7\linewidth]{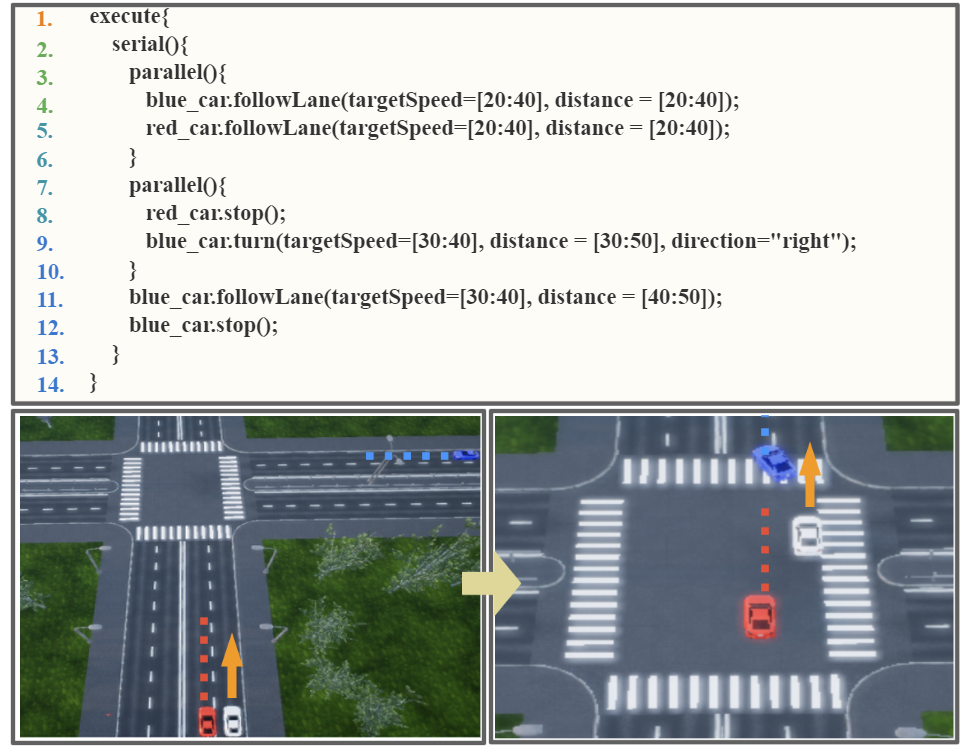}
	\caption{Turning scenario. The ego and red car go straight through an "+" intersection. The blue car follows the lane and then turns right.}
    \label{experiment3}
    %\vspace{-11pt}
\end{figure}

Previous works \cite{ding2020learning,ding2021multimodal,ding2021causalaf} considered various scenarios consisting of only one step. In our implementation, we take the same actions from these scenarios to construct a more complex overtaking scenario with five steps. Besides, we design another turning scenario to show the generalization ability of our model in Fig. \ref{experiment3}.

%
%include ego turning left at an intersection while a cyclist crosses the road, opposite vehicle running red light while ego passes through an intersection, 
% (Opposite Vehicle Running Red Light)
%ego turns left at an intersection while another vehicle approaches from the opposite direction,
% (Unprotected Left Turn)
%and ego turns right at an intersection while another vehicle approaches from the left.
% (Signalized Junction Right Turn)

\subsection{Comparison Experiment}

 \begin{figure}[tbp]
 \centering
     \includegraphics[width=0.6\linewidth]{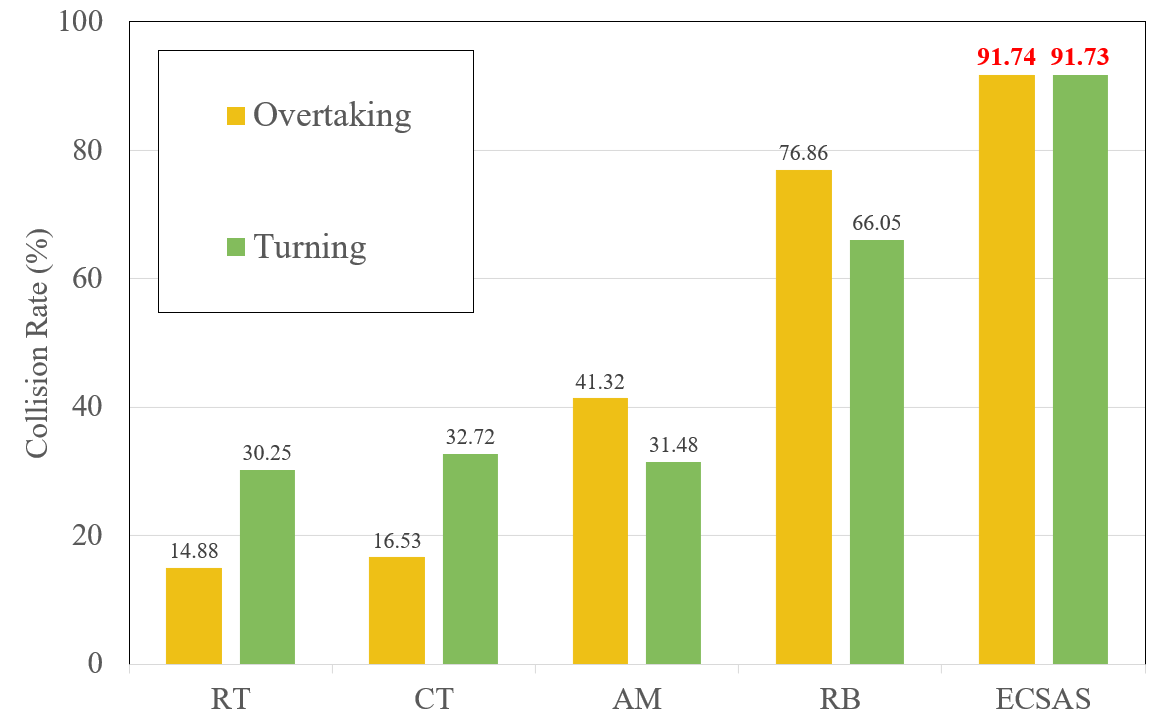}
	\caption{The collision rate of RT, CT, the ablation of action mask(AM) and replay buffer(RB), and ECSAS.  }
    \label{experiment1}
\end{figure}

\begin{figure}[htbp]
 \centering
     \includegraphics[width=\linewidth]{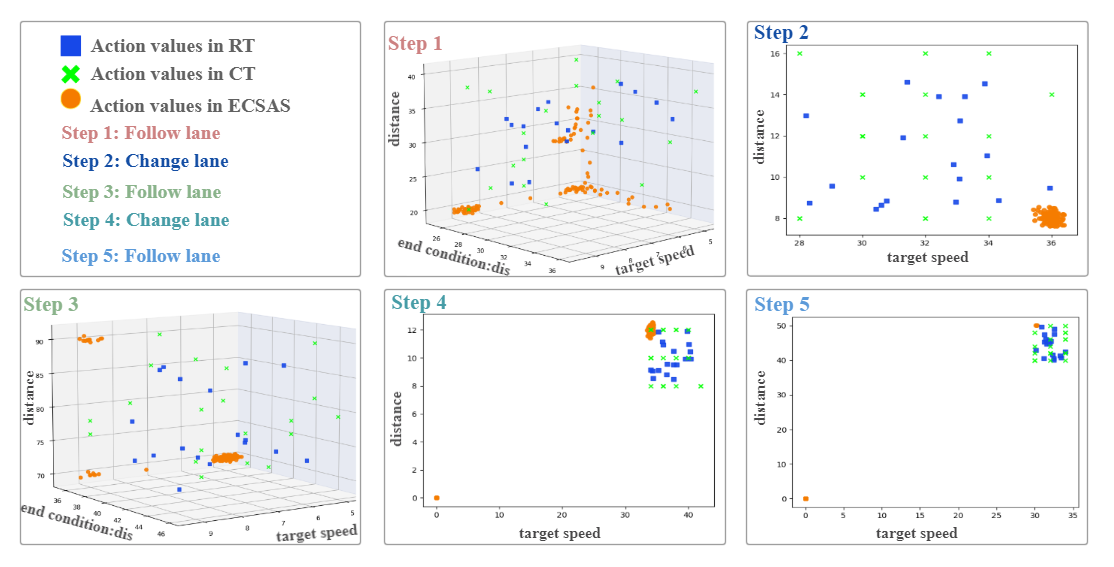}
	\caption{Action value analysis in overtaking scenarios. 
As described in Fig.\ref{framework} Scenarios Representation, in steps 1 and 3, the actions have three parameters, thus are shown in 3D. While in steps 2, 4 and 5, the actions have two parameters, thus are shown in 2D. Orange circle, green cross, and blue squares respectively represent the action values leading to the collision  of ECSAS, CT, and RT in each step of simulation scenarios. For ECSAS, the action values leading to collisions in step 1 are rather discrete. However, in steps 2 and 3, affected by Action Mask in step 1, action values by ECSAS start to gather towards several regions, which means the action sequences are efficient. In each step, there are different action values leading to a collision, indicating the ECSAS can find different collision action distributions of different steps.}    
    \label{experiment2}
    % \vspace{-16pt}
\end{figure}
Fig. \ref{experiment1} compares ECSAS with other methods.
\begin{itemize}
     \item \textit{RT}: We sample all  values  from the specifications of actions by BTScenario.  This method may find critical scenarios but does not consider the influence of the combination of the action parameters.
    \item \textit{CT}: The core is to cover the pairwise discrete combinations of multiple factor values with the fewest test cases. To reduce the searching expense, we discrete each action range, taking values every two numbers by Microsoft PICT tool.  
    \item \textit{Ablation Experiments(AM and RB)}: We evaluate the contributions of the action mask and replay buffer optimization for ECSAS.
\end{itemize}

ECSAS consistently outperforms all the approaches on the collision rate. For example, ECSAS achieved 76.86\% and 75.21\% boost on the metric compared with random search methods (RT and CT) in the overtaking scenario. Since the critical scenario is at the tail of the long-tailed distribution, random search methods are inefficient. By considering the temporal relationship of parameters, ECSAS gains remarkable improvement.

We observe that the ablation results of the action mask and replay buffer optimization decrease compared with the full model in various scenarios, which indicates that the action mask and replay buffer optimization are beneficial for focusing on information about past actions and historical collision scenarios in CSG.

From Fig. \ref{experiment2}, we analyze the action values of collisions in overtaking scenarios, which further verify the impact of action sequences on generating critical scenarios.
The action values leading to collisions  in step 1 are rather discrete. However, in steps 2 and 3, affected by actions in step 1, action values by ECSAS start to gather towards a region. 
Since RL converges the direction with the least loss, tending to a local optimal, it is likely to fall into mode collapse where the generation of action values is aggregated in one action space. In future work, we will address this issue.
Moreover, since vehicles collide during the first three steps, some values of speed and distance in step 4 and step 5 become zero -- their values are of no importance anymore.

\section{Conclusion}
\label{sec:conclusion}
In this paper, we propose  ECSAS for critical scenario generation, which focuses on the correlation between parameters of the action sequence in a logical scenario. We model the action sequence by our description language, BTScenario. We adopt RL with the action mask and replay buffer optimization to sample critical parameters. Our model outperforms the RT and CT remarkably in various scenarios.

Some potential improvements can be explored in future work: $\left( 1 \right)$ We will use different RL algorithms to analyze perception, decision, and control algorithms. $\left( 2 \right)$ We will further extend BTScenario 
to support some random actions, random events, and distribution functions for range parameters. 
%
% ---- Bibliography ----
%
% BibTeX users should specify bibliography style 'splncs04'.
% References will then be sorted and formatted in the correct style.
%
% Generated by IEEEtran.bst, version: 1.14 (2015/08/26)

%

\end{document}